# New Probabilistic Bounds on Eigenvalues and Eigenvectors of Random Kernel Matrices

**Nima Reyhani**
Aalto University, School of Science
Helsinki, Finland
nima.reyhani@aalto.fi

**Hideitsu Hino**
Waseda University
Tokyo, Japan
hideitsu.hino@toki.waseda.jp

**Ricardo Vigário**
Aalto University, School of Science
Helsinki, Finand
ricardo.vigario@aalto.fi

## Abstract

Kernel methods are successful approaches for different machine learning problems. This success is mainly rooted in using feature maps and kernel matrices. Some methods rely on the eigenvalues/eigenvectors of the kernel matrix, while for other methods the spectral information can be used to estimate the excess risk. An important question remains on how close the sample eigenvalues/eigenvectors are to the population values. In this paper, we improve earlier results on concentration bounds for eigenvalues of general kernel matrices. For distance and inner product kernel functions, e.g. radial basis functions, we provide new concentration bounds, which are characterized by the eigenvalues of the sample covariance matrix. Meanwhile, the obstacles for sharper bounds are accounted for and partially addressed. As a case study, we derive a concentration inequality for sample kernel target-alignment.

## 1 INTRODUCTION

Kernel methods such as Spectral Clustering, Kernel Principal Component Analysis(KPCA), and Support Vector Machines, are successful approaches in many practical machine learning and data analysis problems (Steinwart & Christmann, 2008). The main ingredient of these methods is the kernel matrix, which is built using the kernel function, evaluated at given sample points. The kernel matrix is a finite sample estimation of the kernel integral operator, which is determined by the kernel function. A number of algorithms rely on the eigenvalues and eigenvectors of the sample kernel operator/kernel matrix, and employ this information for further analysis. KPCA (Zwald & Blanchard, 2006) and kernel target-alignment (Cristianini et. al., 2002) and (Jia & Liao, 2009) are good examples of such procedure. Also, the Rademacher complexity, and therefore the excess loss of the large margin classifiers can be computed using the eigenvalues of the kernel operator, see (Mendelson & Pajor, 2005) and references therein. Thus, it is important to know how reliably the spectrum of the kernel operator can be estimated, using a finite samples kernel matrix.

The importance of kernel and its spectral decomposition has initiated a series of studies of the concentration of the eigenvalues and eigenvectors of the kernel matrix around their expected values. Under certain rate of spectral decay of the kernel integral operator, (Koltchinskii, 1998) and (Koltchinksii & Giné, 2000) showed that the difference between population and sample eigenvalues and eigenvectors of a kernel matrix is asymptotically normal. A non-asymptotic analysis of kernel spectrum was first presented in (Shawe-Taylor et. al., 2005) and recently revisited in (Jia & Liao, 2009), where they derive an exponential bound for the concentration of the sample eigenvalues, using the bounded difference inequality. Also, using concentration results for isoperimetric random vectors, (Mendelson & Pajor, 2005) derived a concentration bound for the supremum for the concentration of eigenvalues. The bound is determined by the Orlicz norm of the kernel function with respect to the data distribution, the number of dimensions and the number of samples.

In this paper, we establish sharper exponential bounds for the concentration of sample eigenvalues of general kernel matrices, which depends on the minimum distance between a given eigenvalue and the rest of the spectrum. Separately, for Euclidean-distance and inner-product kernels, we provide a set of different bounds, which connect the concentration of eigenvalues to the maximum spectral gap of the sample covariance matrix. Similar results are derived for the eigenvectors of the same kernel matrix. Some experiments are also designed, to empirically test the presented re-sults. As a case study, we derive concentration bounds for kernel target-alignment, which can be used to measure the agreement between a kernel matrix and the given labels.

The paper is organized as follows. Section 2 summarizes the previous results on concentration bounds for eigenvalues. In section 3, we present new concentration bounds for eigenvalues and eigenvectors of kernel matrices. Section 4 provides a concentration inequality for sample kernel alignment as a case study.

## 2 PREVIOUS RESULTS

In this section, we briefly present some of the most relevant earlier results on the concentration of the kernel matrix eigenvalues. The main assumption followed throughout the paper is summarized as

**Assumption 1.** *Let $\mathcal{S} = \{\bm{x}_1, \ldots, \bm{x}_n\}$, $\bm{x}_i \in \mathcal{X} \subseteq \mathbb{R}^p$, be a set of samples of size $n$, independently drawn from distribution $P$. Let us define $K_n \in \mathbb{R}^{n \times n}$ to be a kernel matrix on $\mathcal{S}$, with $[K_n]_{i,j} = \frac{1}{n} k(\bm{x}_i, \bm{x}_j)$, for Mercer's kernel function $k$.*

The kernel function $k$ defines a kernel integral operator by

$$Tf(\cdot) = \int k(\bm{x}, \cdot) f(\bm{x}) dP(\bm{x}),$$

for all smooth functions $f$. The eigenvalue equation is defined by $T\bm{u}(\cdot) = \lambda \bm{u}(\cdot)$, where $\lambda$ is an eigenvalue of $T$ and $\bm{u} : \mathbb{R}^p \to \mathbb{R}$ is the corresponding eigenfunction. For data samples $\mathcal{S}$, the kernel operator can be estimated by the kernel matrix $K_n$. Then, the eigenvalue problem can be reduced to

$$K_n \bm{u} = \lambda \bm{u}.$$

The solutions of the above equation are denoted by $\lambda_i(K_n)$ and $\bm{u}_i(K_n)$, $i = 1, \ldots, n$. For simplicity of the paper, we assume the eigenvalues of kernel operator/matrix are not identical and are indexed in decreasing order, i.e. $\lambda_1 > \lambda_2 > \ldots$. The focus of this paper is mainly in finding bounds for the concentrations of type: $P\{|\frac{1}{n}\lambda_i(K_n) - \frac{1}{n}\mathbb{E}_\mathcal{S} \lambda_i(K_n)| \leq \epsilon\}$ and similarly $P\{\|\frac{1}{n}\bm{u}_i(K_n) - \frac{1}{n}\mathbb{E}_\mathcal{S} \bm{u}_i(K_n)\| \leq \epsilon\}$, for all $i = 1, \ldots, n$. We denote the finite dimensional unit sphere in $\mathbb{R}^p$ by $\mathcal{S}^{p-1}$, i.e. $\mathcal{S}^{p-1} := \{\bm{w} : \bm{w} \in \mathbb{R}^p, \|\bm{w}\|_2 = 1\}$, and the Lipschitz norm of function $f$ by $|f|_L$.

The following theorem is presented in (Shawe-Taylor et. al., 2005) and provides a uniform bound that depends only on the supremum of the diagonal elements of the kernel matrix. This theorem requires that the kernel is bounded.

**Theorem 2.1** ((Shawe-Taylor et. al. 2005) ). *Let us take Assumption 1. Then, for all $\epsilon > 0$, we have*

$$P\{|\frac{1}{n}\lambda_i(K_n) - \mathbb{E}_\mathcal{S} \frac{1}{n}\lambda_i(K_n)| > \epsilon\} \leq 2\exp(-\frac{2n\epsilon^2}{R^4}),$$

*where $R^2 = \max_{\bm{x} \in \mathbb{R}^p} k(\bm{x}, \bm{x})$.*

The bound provided by Theorem 2.1 is suboptimal, as it only depends on the kernel function $k$ through $R$ and also it is uniform for all different eigenvalues. To address the suboptimality of Theorem 2.1, (Jia & Liao, 2009) proposed to bound the concentration of the kernel matrix eigenvalues using the largest eigenvalue of the sample kernel and a quantity $\theta \in (0, 1)$:

**Theorem 2.2** ((Jia & Liao, 2009)). *Let us take Assumption 1. Then, for every $\epsilon > 0$, there exists $0 < \theta \leq 1$, such that*

$$P\{|\frac{1}{n}\lambda_i(K_n) - \mathbb{E}_\mathcal{S} \frac{1}{n}\lambda_i(K_n)| > \epsilon\} \leq 2\exp(-\frac{2\epsilon^2}{\theta^2 \lambda_1^2(K_n)}).$$

The advantage of this result over the uniform result in Theorem 2.1 is that the concentration bound depends solely on a parameter $\theta$ and the largest sample eigenvalue of the kernel matrix. The parameter $\theta$, in principle, varies as a function of the order of the eigenvalue. However, Theorem 2.2 does not provide any hint on estimating an optimal value for the parameter $\theta$ for any specific eigenvalue. In the following section, we provide sharper concentration inequalities for the spectral decomposition of the kernel matrix.

## 3 RESULTS

The main result is presented in Theorem 3.1, which establishes a bound on the concentration of eigenvalues of the kernel matrix, for any general kernel function that fulfills Mercer's theorem. In the next subsections, we provide concentration bounds for the spectral decomposition of the distance and inner product kernels. Our first result is inspired by the following observation: we built kernel matrices, using Gaussian kernels from 100 samples drawn from a 5-dimensional multivariate Gaussian. We repeat this example 1000 times. The boxplot of the first 15 eigenvalues of the Gaussian kernel matrix are depicted in Figure 1. From the boxplot, we can see that the distance between any box's corner and location of median decreases as the gap between corresponding eigenvalue and the spectrum of the kernel matrix decreases. This suggests that the concentration of the eigenvalues might be controlled by the distance of that eigenvalue to the spcetrum. None of the earlier results, i.e. Theorems 2.1 and 2.2, provide such a characterization. Theorem 3.1 establishes a concentration bound, which illustrates the aforementioned phenomenon.

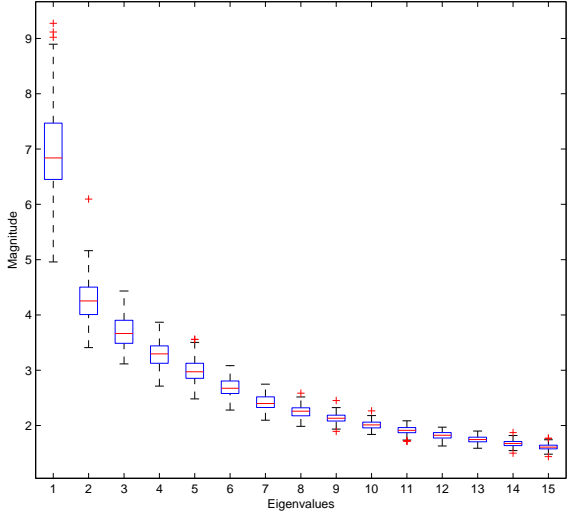

Figure 1: Boxplot of 15 first eigenvalues of Gaussian kernels with 5 dimensional Gaussian samples, drawn from $\mathcal{N}(0, I_5)$.

**Theorem 3.1** (Probabilistic bound on eigenvalues of kernel matrices). *Let us take (Assumption 1) and define $\lambda_{i,i+1}(K_n) := \lambda_i(K_n) - \lambda_{i+1}(K_n)$. Then, the following inequality holds for sample eigenvalues, $\lambda_i(K_n)$, $i = 1, \ldots, n-1$.*

$$P\{|\frac{1}{n}\lambda_i(K_n) - \mathbb{E}_{\mathcal{S}}\frac{1}{n}\lambda_i(K_n)| > \epsilon\} \leq \exp(-\frac{2n\epsilon^2}{\lambda_{i,i+1}^2(K_n)}). \quad (1)$$

Experimental results to validate the above bound are given in Example 1. To prove Theorem 3.1, we introduce Cauchy's interlacing lemma, as a particular matrix perturbation result and bounded difference inequality.

**Definition 1** (Principal Submatrix). *The principal submatrix of A of order $n - 1$ is the matrix obtained by taking the first $n - 1$ rows and columns of matrix A.*

**Lemma** (Cauchy's Interlacing lemma, (Horn & Johnson, 1985 )). *Let B be a principal submatrix of the Hermitian matrix A, of order $n-1$, with eigenvalues $\mu_1 \geq \cdots \geq \mu_{n-1}$. Also, we denote the eigenvalues of A by $\lambda_1 \geq \cdots \geq \lambda_n$. Then, the following property holds for eigenvalues of A and B:*

$$\lambda_1 \geq \mu_1 \geq \lambda_2 \geq \mu_2 \geq \cdots \geq \mu_{n-1} \geq \lambda_n.$$

**Theorem** (Bounded Difference Inequality, (Ledoux & Talagrand, 1991) and (Shawe-Taylor et. al., 2005)). *Let $\boldsymbol{x}_1, \ldots, \boldsymbol{x}_n$ be independent random variables taking values in a set $\mathcal{A}$ and $g : \mathcal{A}^n \to \mathbb{R}$ satisfies one of the following: for $c_i \geq 0, 1 \leq i \leq n$,*

$$\sup_{x_1,\ldots,x_n,x_i'} |g(x_1,\ldots,x_n) - g(x_1,\ldots,x_i',x_{i+1},x_n)| \leq c_i$$

$$\sup_{x_1,\ldots,x_n} |g(x_1,\ldots,x_n) - g_i(x_1,\ldots,x_{i-1},x_{i+1},x_n)| \leq c_i,$$

*where $\boldsymbol{x}_1', \ldots, \boldsymbol{x}_n'$ are independent copies of $\boldsymbol{x}_1, \ldots, \boldsymbol{x}_n$ and $g_i : \mathcal{A}^{n-1} \to \mathbb{R}, i = 1, \ldots, n$. Then, for all $\epsilon > 0$ we have $P\{|g(\boldsymbol{x}_1, \ldots, \boldsymbol{x}_n) - \mathbb{E}g(\boldsymbol{x}_1, \ldots, \boldsymbol{x}_n)| > \epsilon\} \leq \exp(-\frac{2\epsilon^2}{\sum_{i=1}^n c_i^2})$.*

**Proof of Theorem 3.1.** Let us define submatrix $K_n^n \in \mathbb{R}^{n-1 \times n-1}$ with $[K_n^n]_{i,j} = k(\boldsymbol{x}_i, \boldsymbol{x}_j), i, j = 1, \ldots, n - 1$. By the Cauchy's interlacing lemma, we have $\lambda_i(K_n) - \lambda_i(K_n^n) \leq \lambda_i(K_n) - \lambda_{i+1}(K_n) = \lambda_{i,i+1}(K_n)$. Using the bounded difference inequality, see Appendix 1, and taking $\sum_{j=1}^n c_j^2 = n\lambda_{i,i+1}^2(K_n)$ we obtain the claimed result. □

Note that, $\lambda_{i,i+1}(K_n)$ can be bounded by $\theta\lambda_1$ for some $\theta \in (0, 1]$, which reduces to the bound in Theorem 2.2. With extra conditions on the condition number of kernel matrix we can derive the bound in Theorem 2.1 by the above result, for details please see Corollary 2 in (Jia & Liao, 2009).

**Corollary 3.1.** *Using the assumptions of Theorem 3.1, the sum of the top and low eigenvalues are concentrated around their mean, as follows:*

$$P\{|\frac{1}{n}\sum_{i=1}^k \lambda_i(K_n) - \mathbb{E}_S \frac{1}{n}\sum_{i=1}^k \lambda_i(K_n)| > \epsilon\}$$
$$\leq \exp(-\frac{2n\epsilon^2}{\lambda_{1,k+1}^2(K_n)}),$$
$$P\{|\frac{1}{n}\sum_{i=k}^n \lambda_i(K_n) - \mathbb{E}\frac{1}{n}\sum_{i=k}^n \lambda_i(K_n)| > \epsilon\}$$
$$\leq \exp(-\frac{2n\epsilon^2}{\lambda_{k,n}^2(K)}),$$

*where $\lambda_{1,k+1}(K_n) = \lambda_1(K_n) - \lambda_{k+1}(K_n)$ and $\lambda_{k,n}(K_n) = \lambda_k(K_n) - \lambda_n(K_n)$.*

The above inequality is useful to derive a probabilistic bound for the population risk of Kernel PCA, using the empirical risk. We leave further details for future work.

### 3.1 DISTANCE AND INNER PRODUCT KERNELS

In the following, we present results for distance and inner product kernel functions, which are commonly used in practical data analysis, such as Radial Basis Functions in SVM (Steinwart & Christmann, 2008), or Laplacian kernels in Spectral Clustering (Bengio et.

al., 2003). The distance kernel matrix and inner product kernel matrix are defined by $k(\boldsymbol{x}_i, \boldsymbol{x}_j) = f(\|\boldsymbol{x}_i - \boldsymbol{x}_j\|_2^2)$ and $k(\boldsymbol{x}_i, \boldsymbol{x}_j) = f(\boldsymbol{x}_i^\top \boldsymbol{x}_j)$, where $f : \mathbb{R} \to \mathbb{R}$ is a smooth function, which has a Bochner's integral representation (Steinwart & Christmann, 2008). For the rest of the paper, except section 4, the spectral norm is denoted by $\|\cdot\|$. The results for concentration of eigenvalues of the distance kernel matrix is presented in the Theorem 3.2.

**Theorem 3.2.** *Let us take (Assumption 1), with $k(\boldsymbol{x}_i, \boldsymbol{x}_j) = f(\|\boldsymbol{x}_i - \boldsymbol{x}_j\|_2^2)$. We assume $\boldsymbol{x}_1, \ldots, \boldsymbol{x}_n$ have a nonsingular sample covariance matrix $\Sigma$. Also we assume, for r.v.s $\boldsymbol{y}_1, \ldots, \boldsymbol{y}_n$ with identity covariance matrix and same distribution as $\boldsymbol{x}_1, \ldots, \boldsymbol{x}_n$, $\|\boldsymbol{y}_i\|_2 \leq M$ and $\boldsymbol{x}_i = \Sigma^{\frac{1}{2}} \boldsymbol{y}_i, \forall i = 1, \ldots n$ holds. The first and last eigenvalues of $\Sigma$ are denoted by $\lambda_1(\Sigma)$ and $\lambda_p(\Sigma)$, respectively. Then, for $\lambda_{1,p} = \lambda_1(\Sigma) - \lambda_p(\Sigma)$, we have*

$$P\{|\frac{1}{n}\lambda_i(K) - \mathbb{E}_\mathcal{S} \frac{1}{n}\lambda_i(K)| > \epsilon\} \leq$$
$$\exp(-\frac{n^2 \epsilon^2}{18 M^4 |f|_L^2 \lambda_{1,p}^2}). \quad (2)$$

**Proof.** Let us denote the perturbation of matrix $K$ by $K^n$ defined by $K_{i,j}^n := \frac{1}{n} f(\|\boldsymbol{x}_i - \boldsymbol{x}_j\|_2^2), \forall i, j = 2, \ldots, n$ and $K_{j,1}^n = K_{1,j}^n := \frac{1}{n} f(\|\boldsymbol{x}_{1'} - \boldsymbol{x}_j\|), \forall j = 1', \ldots, n$, where $\boldsymbol{x}_{1'}$ is independent copy of $\boldsymbol{x}_i, \forall i = 1, \ldots, n$. Then, $K = K^n + E$. Suppose the perturbation term $E$ has a small spectral norm. Then, by the Wielandt Theorem (Horn & Johnson, 1985), we have $\left|\frac{1}{n}\lambda_i(K_n) - \frac{1}{n}\lambda_i(K_n^n)\right| \leq \frac{1}{n}\|E\|$. The operator norm of $E$ can be computed as follows.

$\|E\|$
$= \|K_n - K_n^n\| = \sup_{\boldsymbol{z} \in \mathcal{S}^{n-1}} |\langle (K_n - K_n^n)\boldsymbol{z}, \boldsymbol{z}\rangle|$
$= \frac{2}{n} \sup_{\boldsymbol{z} \in \mathcal{S}^{n-1}} \left|\sum_{i=1}^n z_1 z_i [f(\|\boldsymbol{x}_i - \boldsymbol{x}_1\|_2^2) - f(\|\boldsymbol{x}_i - \boldsymbol{x}_{1'}\|_2^2)]\right|$
$\leq \frac{2}{n} \sup_{\boldsymbol{z} \in \mathcal{S}^{n-1}} |z_1| \left|\sum_{i=1}^n z_i [f(\|\boldsymbol{x}_i - \boldsymbol{x}_1\|_2^2) - f(\|\boldsymbol{x}_i - \boldsymbol{x}_{1'}\|_2^2)]\right|$
$\leq \frac{2}{n} \sup_{\boldsymbol{z} \in \mathcal{S}^{n-1}} \left(\sum_{i=1}^n z_i^2\right)^{\frac{1}{2}}$
$\cdot \left(\sum_{i=1}^n [f(\|\boldsymbol{x}_i - \boldsymbol{x}_1\|_2^2) - f(\|\boldsymbol{x}_i - \boldsymbol{x}_{1'}\|_2^2)]^2\right)^{\frac{1}{2}}$
$\leq \frac{2|f|_L}{n} \left(\sum_{i=1}^n \left|\|\boldsymbol{x}_i - \boldsymbol{x}_1\|_2^2 - \|\boldsymbol{x}_i - \boldsymbol{x}_{1'}\|_2^2\right|^2\right)^{\frac{1}{2}},$

where we used Hölder's inequality. Now, by the bounded Manifold assumption, i.e. $\|\boldsymbol{y}_i\| \leq M$, we have,

$|\|\boldsymbol{x}_i - \boldsymbol{x}_1\|_2^2 - \|\boldsymbol{x}_i - \boldsymbol{x}_{1'}\|_2^2|$
$= |\boldsymbol{x}_1^\top \boldsymbol{x}_1 - \boldsymbol{x}_{1'}^\top \boldsymbol{x}_{1'} - 2\boldsymbol{x}_1^\top \boldsymbol{x}_i + 2\boldsymbol{x}_{1'}^\top \boldsymbol{x}_i|$
$= |\boldsymbol{y}_1^\top \Sigma \boldsymbol{y}_1 + 2\boldsymbol{y}_{1'}^\top \Sigma \boldsymbol{y}_i - \boldsymbol{y}_{1'}^\top \Sigma \boldsymbol{y}_{1'} - 2\boldsymbol{y}_1^\top \Sigma \boldsymbol{y}_i|$
$\leq 3M^2 \lambda_{1,p}(\Sigma).$

Therefore, the spectral norm of the error matrix can be bounded by,

$$\|E\| \leq \frac{6M^2}{\sqrt{n}} |f|_L \lambda_{1,p}(\Sigma). \quad (3)$$

By the bounded difference inequality and eq. 3 we obtain the desired result. □

Similar results to Corollary 3.1 can be derived, where the kernel eigenvalues in the exponential bound are replaced by eigenvalues of the sample covariance matrix, in the same way as they appear in the exponential bound of Theorem 3.2. The concentration bounds for the inner product kernel can be derived in a similar way as in Theorem 3.2. The sketch of the derivation is provided in Remark 3.1.

**Remark 3.1** (concentration bound for smooth inner product kernels). *For the inner product kernel as defined earlier in this section with $|f|_L$-Lipschitz function we have $P\left\{\left|\frac{1}{n}\lambda_i(K_n) - \mathbb{E}_\mathcal{S} \frac{1}{n}\lambda_i(K_n)\right| > \epsilon\right\} \leq \exp\left\{-\frac{n^2 \epsilon^2}{4|f|_L^2 M^4 \lambda_{1,p}^2(\Sigma)}\right\}$. The proof is similar to as of Theorem 3.2.* ◁

Theorem 3.2 and the result in Remark 3.1 connects the concentration of the eigenvalues of the distance and inner product kernel matrices to the difference between the largest and smallest eigenvalues of the covariance matrix. In high dimensional data, the first eigenvalue of the sample covariance matrix is likely to become rather large, implying that the eigenvalues of the distance kernel matrix are not concentrated around the mean. This result suggests that it may be better not to use smooth distance, or inner product kernels in high dimensional data. The results of this section hold also for a wider class of kernel functions that behaves almost like the distance or the inner product kernels. The characterization of the wider class of this functions is summarized in Remark 3.2.

**Remark 3.2.** *The result of theorem 3.2 also holds for other kernels that satisfy one of the following conditions.*

1. $|k(\boldsymbol{x}_1, \boldsymbol{x}_2) - k(\boldsymbol{x}_1', \boldsymbol{x}_2')|$
$\leq C_\mathcal{X} \left|\|\boldsymbol{x}_1 - \boldsymbol{x}_1'\|_2^2 - \|\boldsymbol{x}_2 - \boldsymbol{x}_2'\|_2^2\right|,$
2. $|k(\boldsymbol{x}_1, \boldsymbol{x}_2) - k(\boldsymbol{x}_1', \boldsymbol{x}_2')| \leq C_\mathcal{X}' \left|\boldsymbol{x}_1^\top \boldsymbol{x}_1' - \boldsymbol{x}_2^\top \boldsymbol{x}_2'\right|,$

*where $C_\mathcal{X}$ and $C_\mathcal{X}'$ are constants depending on the smoothness of the kernel function and the data distribution.* ◁

In order to check the bounds in Theorems 3.1 and 3.2, we run two numerical experiments with Gaussian kernels. The details and results are summarized in Example 1.

**Example 1.** *We draw 100 samples from the standard normal distribution, and compute the kernel matrix using a Gaussian kernel function $k(x,y) = \exp(-0.5\|x-y\|_2^2)$. We repeat this procedure a 1000 times, and compute the empirical estimate of the left hand and right hand sides of inequality (1). The results are depicted in Figure 2 (top), where solid, dashed, and dotted lines correspond to left hand side of eq. (1) for $i = 1,2,3$, respectively. Lines with different marks correspond to right hand side of inequality (1) for $i = 1,2,3$, respectively. As can be seen, the concentration of eigenvalues varies by their order. The bound presented in eq. (1) tracks the concentration changes for each eigenvalue separately.*

*Similarly we repeat the above example for multivariate Gaussian samples $\mathcal{N}(0, I_2)$ and $\mathcal{N}(0, I_5)$, to check the bound in Theorem 3.2. $I_p$ denotes the identity matrix in $\mathbb{R}^p$. The empirical estimate of left hand side and right hand side of inequality (2) are drawn in Figure 2 (bottom). As the experiments illustrate, the concentration of eigenvalues change with the dimension of samples, which can be captured by the bound provided in inequality (2).*

One of the main obstacles to improve the bounds derived either in this or earlier section is due to the limitation of the matrix perturbation results. For example, the results in Theorem 3.2 and Theorem 2.1 rely on the first order eigenvalue expansion $\lambda_i(\tilde{K}) = \lambda_i(K) - \boldsymbol{u}_i^\top(K)E\boldsymbol{u}_i(K) + O(\|K - \tilde{K}\|^2)$, where $\tilde{K}$ is a perturbation of the matrix or compact linear operator $K$ and $E = \tilde{K} - K$. However, these results can be slightly improved by using the second order expansion, as described in (Kato, 1996):

$$\lambda_i(\tilde{K}) = \lambda_i(K) - \boldsymbol{u}_i^\top(K)E\boldsymbol{u}_i(K)$$
$$+ \sum_{\substack{j=1 \\ j \neq i}}^n \frac{u_j^\top(K)E\boldsymbol{u}_i(K)u_i^\top(K)E\boldsymbol{u}_j(K)}{\lambda_j - \lambda_i} + O(\|E\|^3).$$

The above expansion holds if the spectral norm of the error operator/matrix $E$ is smaller than half of the distance between the eigenvalue $\lambda_i$ and the rest of the spectrum of $K$. In the notation of Theorem 3.2, $E = K_n - K_n^n$ is symmetric. Therefore, the perturbation of the eigenvalues of the kernel matrix can be bounded by

$$|\lambda_i(K_n) - \lambda_i(K_n^n)| \leq \|E\|_2 + \|E\|_2^2 \sum_{\substack{j=1 \\ j \neq i}}^n \frac{1}{|\lambda_j - \lambda_i|}. \quad (4)$$

Combining the preceding bound with eq. (3) reads

$$P\left\{\left|\frac{1}{n}\lambda_i(K_n) - \mathbb{E}_{\mathcal{S}}\frac{1}{n}\lambda_i(K_n)\right| > \epsilon\right\} \leq \exp\left\{-\frac{n^2\epsilon^2}{\gamma^2}\right\},$$

where

$$\gamma = 6M^2|f|_L \frac{\lambda_{1,p}(\Sigma)}{\sqrt{n}} + 36M^4|f|_L^2 \frac{\lambda_{1,p}^2(\Sigma)}{n} \sum_{\substack{j=1 \\ j \neq i}}^n \frac{1}{\lambda_{j,i}^2(K_n)},$$

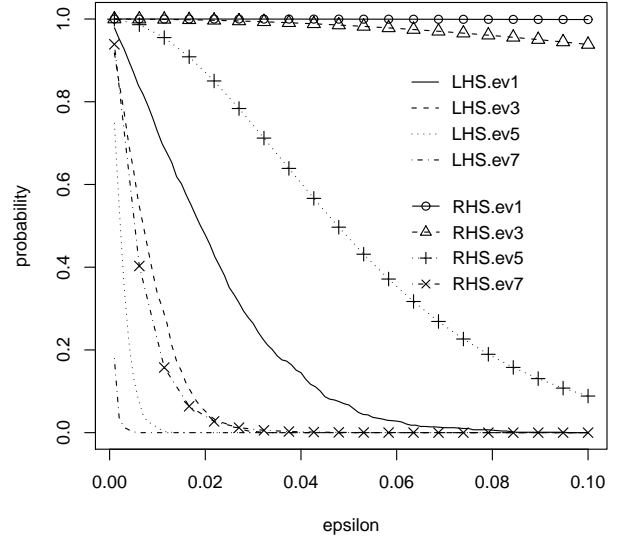

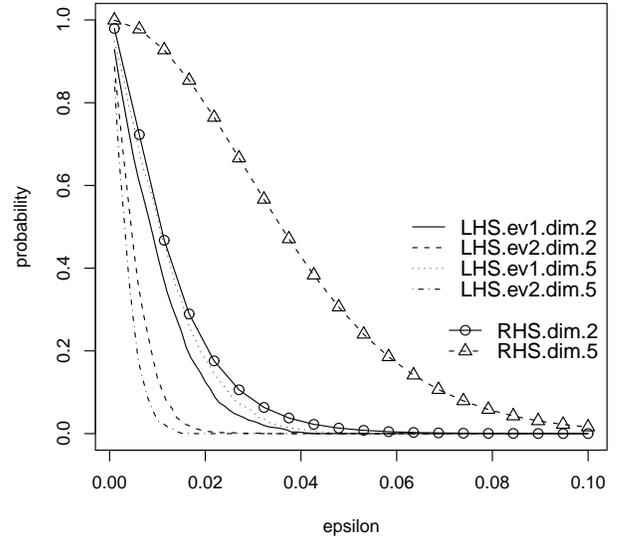

Figure 2: Empirical study on concentrations of eigenvalues. The top figure corresponds to inequality (1) for 1st, 3rd, and 5th eigenvalues. The bottom figure corresponds to inequality (2). See Example 1 for details.

and $\lambda_{j,i}(K_n) = \lambda_j(K_n) - \lambda_i(K_n)$.

Another direction to improve this result is to improve the computation of the spectral norm $\|K_n - K_n^n\|$, so that the dependencies between elements of $E$ are taken into account. This is left for future work.

## 3.2 EIGENVECTORS OF INNER PRODUCT AND DISTANCE KERNELS

In this section, we provide the concentration bounds for the eigenvectors of the distance and inner product kernel matrices. The results mainly rely on the eigenfunction perturbation expansion summarized in the following theorem, which is due to (Kato, 1996).

**Theorem 3.3.** *[eigenfunction expansion] Let $\tilde{K} = K + E$, and assume $\|E\|$ is smaller than half of the distance between eigenvalue $\lambda_i$ and the rest of the eigenvalues, then*

$$\tilde{u}_i = u_i + \sum_{\substack{j=1 \\ j \neq i}}^n \frac{u_j^\top E u_i}{\lambda_j - \lambda_i} u_j + O(\|E\|^2), \quad (5)$$

*where $u_i$ is the $i$-th eigenvector of matrix $K$, with corresponding eigenvalue $\lambda_i$, and $\tilde{u}$ is the eigenvector of $\tilde{K}$ corresponding to $\lambda_i$.*

Note that, by eq. (5), when $|\lambda_j - \lambda_i|$ is small, the term $\frac{E}{\lambda_j - \lambda_i}$ has a large norm, which results in an instability of eigenvector $u_i$ due to the perturbation. This is the case for all eigenvectors that correspond to small eigenvalues. Such phenomenon is empirically studied in (Ng et. al., 2001). In the following, we provide pointwise and uniform bounds for the concentration of eigenvectors using eq. (5).

The following results are slightly different from those in (Zwald & Blanchard, 2006), as it contain one extra projection into the given samples. Moreover, the proof presented here holds also for kernels with an infinite dimensional feature space. In the following we denote the evaluation of function $u(\cdot) : \mathbb{R}^p \to \mathbb{R}$ on samples $\mathcal{S}$ by $u|_\mathcal{S} := [u(x_1), \ldots, u(x_n)]$.

**Lemma 1.** *Let us take the assumptions made in Theorem 3.2. We consider a real valued function $f_w(\cdot) : \mathbb{R}^n \to \mathbb{R}, x \mapsto \langle \cdot, w \rangle, w \in \mathbb{R}^n$. Further, assume $k(x, y) = f(\|x - y\|_2^2)$. Then, for every $w \in \mathcal{S}^{n-1}$ we have the following concentration result:*

$$P\{|f_w(u_i(K_n)) - \mathbb{E}_\mathcal{S} f_w(u_i(K_n)|_\mathcal{S})| > \epsilon\}$$
$$\leq \exp\left(-\frac{\epsilon^2}{18M^4|f|_L^2 R_i^2(K_n)\lambda_{1,p}^2(\Sigma)}\right),$$

*where $u_i(K_n)$ is the eigenvector of kernel matrix $K_n$ corresponding to $i$-th eigenvalue, $\lambda_{1,p}(\Sigma) := \lambda_1(\Sigma) - \lambda_p(\Sigma)$, and $R_i(K_n) := \sum_{\substack{j=1 \\ j \neq i}}^n \frac{1}{|\lambda_i(K_n) - \lambda_j(K_n)|}$.*

**Proof.** Let us define the perturbed matrix $K_n^n$ by $[K_n^n]_{i,j} = \frac{1}{n}k(x_i, x_j), \forall i, j = 2, \ldots, n$ and $[K_n^n]_{1,j} = K_{j,1}^n = k(x_1', x_j)$. Therefore, $K_n = K_n^n + E$. By definition of $f_w$ we have,

$$|f_w(u_i(K_n)) - f_w(u_i(K_n^n))| = |f_w(u_i(K_n) - u_i(K_n^n))|$$
$$\leq \|w\|\|u_i(K_n) - u_i(K_n^n)\|,$$

where $u_i(K_n)$ and $u_i(K_n^n)$ are eigenvectors of matrices $K_n$ and $K_n^n$. Now, we need to compute the quantity $\|u_i(K_n) - u_i(K_n^n)\|$. Using the perturbation result of Theorem 3.3, we have $\|u_i(K_n) - u_i(K_n^n)\| \leq \|E\| R_i(K_n)$. Then, for all $w \in \mathcal{S}^{n-1}$, applying the bounded difference inequality result in the following concentration bound.

$$P\{|\langle w, u_i(K_n) - \mathbb{E}_\mathcal{S} u_i(K_n)\rangle| > \epsilon\}$$
$$\leq \exp\left(-\frac{2\epsilon^2}{n\|E\|^2 R_i^2(K_n) \sup_w \|w\|_2}\right). \quad (6)$$

By plugging eq. (3) into eq. (6) we obtain the claimed result. $\square$

The concentration bound in Lemma 1 can be extended to real-valued convex Lipschitz functions. An immediate application of Lemma 1 is to derive a uniform concentration bound for the kernel matrix eigenvectors. This result is provided in Corollary 3.2.

**Corollary 3.2** (uniform concentration of kernel eigenvectors). *With the same assumptions made in Theorem 1, we have*

$$P\{\|u_i(K_n) - \mathbb{E}_\mathcal{S} u_i(K_n)|_\mathcal{S}\| > \epsilon\} \leq 2\exp\left(2n - c\epsilon^2\right),$$

*where $c^{-1} = 18M^4|f|_L^2 R_i^2(K_n)\lambda_{1,p}^2(\Sigma)$.*

In the above bound, the positive first term inside the exponential function does not bring any harm, since the first eigenvalue of the sample covariance matrix is of order $O(\sqrt{n})$, which cancels out the effect of the $2n$ term inside the exponential function. The proof relies on Lemma 1 and the $\varepsilon$-Net argument (Ledoux & Talagrand, 1991).

**Proof of Corollary 3.2.** Let us denote the $\varepsilon$-Net of the compact unit sphere $\mathcal{S}^{n-1}$ by $\mathcal{N}_\varepsilon$, for some $\varepsilon \in (0,1)$. Let $x \in \mathcal{S}^{n-1}$ such that $\|u\| = \langle x, u \rangle$, where $u := u_i(K_n)$. Now, we can choose $y \in \mathcal{N}_\varepsilon$ so that $\|x - y\| \leq \varepsilon$ holds. Then, we have $|\langle x, u \rangle - \langle y, u \rangle| \leq \|u\|\|x - y\| \leq \varepsilon\|u\|$. By the triangle inequality, we have $|\langle y, u \rangle| \geq |\langle x, u \rangle| - |\langle x, u \rangle - \langle y, u \rangle| \geq \|u\| - \varepsilon\|u\|$. Putting all together, we obtain

$$\max_{y \in \mathcal{N}_\epsilon} |\langle y, u \rangle| \leq \|u\| \leq \frac{1}{1 - \varepsilon} \max_{y \in \mathcal{N}_\varepsilon} |\langle u, y \rangle|.$$

By using the upper bound in the preceding inequality, we have

$$P\{\|u - \mathbb{E}_S u|_S\| > \epsilon\}$$
$$= P\left\{\sup_{w \in S^{n-1}} |\langle w, u - \mathbb{E}_S u|_S\rangle| > \epsilon\right\}$$
$$\leq P\left\{\frac{1}{1-\varepsilon} \max_{w \in \mathcal{N}_\varepsilon} |\langle w, u - \mathbb{E}_S u|_S\rangle| > \epsilon\right\}$$
$$\leq |\mathcal{N}_\varepsilon| P\{|\langle w, u - \mathbb{E}_S u|_S\rangle| > \epsilon\}$$
$$\leq 2|\mathcal{N}_\varepsilon| \exp(-c\epsilon^2),$$

where in the second line we have used the union bound. By Lemma 9.5. in (Ledoux & Talagrand, 1991), for $\varepsilon = \frac{1}{2}$ we have $|\mathcal{N}_\varepsilon| \leq 6^n$. Putting all together we get the claimed result.

□

The result on the eigenvectors of distance kernel can be applied to a broader class of kernels that satisfies Lipschitz smoothness, see Remark 3.2. In particular, for the inner product kernels, we have $\|E\| \leq 2M^2 \frac{|f|_L \lambda_{1,p}(\Sigma)}{\sqrt{n}}$. Plugging this into inequality (6), we get the desired bounds.

The concentration bounds presented in Theorem 1 and Corollary 3.2 can also be slightly improved by using a higher order expansion of the operator perturbation. For more details on higher order expansions of eigenvectors see (Kato, 1996).

## 4 CASE STUDY: KERNEL TARGET-ALIGNMENT

In this section, we provide an example of using the ingredients of the proof in Section 3, for deriving a concentration inequality for the sample kernel alignment. This is a good example to show how the combination of a proper matrix perturbation result and the concentration inequality provides an informative, simple and easy to compute concentration bound.

Kernel target-alignment is proposed in (Cristianini et. al., 2002), for measuring the agreement between a kernel matrix and the given learning task, i.e. classification. By changing the eigenvalues of the kernel matrix, one can improve the alignment to the labels, i.e. target values, which results in improvement of the learning performance. This approach has been proposed for kernel learning (for more detail see (Cristianini et. al., 2002)). (Jia & Liao, 2009) proposed to use the bound in Theorem 2.2 to derive a concentration inequality for sample kernel alignment. Suppose a sample set $\{(x_i, y_i)\}_{i=1}^n$, $x_i \in \mathbb{R}^p$ and $y_i \in \{-1, +1\}$ is given. Let $K_n \in \mathbb{R}^{n \times n}$ be a Mercer kernel matrix defined by $[K_n]_{i,j} = k(x_i, x_j), \forall 1 \leq i, j \leq n$. The kernel target-alignment or kernel alignment is defined by

$$A(y) = \frac{\langle y \otimes y, k\rangle}{\|y \otimes y\|\|k\|} = \frac{\langle y \otimes y, k\rangle}{\|k\|},$$

where $\langle f, g\rangle = \int_{\mathcal{X}^2} f(x,z)g(x,z)dP(x)dP(z)$. Similarly, the sample kernel target alignment of $K_n$ is defined by

$$A(K_n) = \frac{\langle K_n, Y \otimes Y\rangle_F}{\sqrt{\langle Y \otimes Y, Y \otimes Y\rangle_F \langle K, K_n\rangle}} = \frac{Y^\top K_n Y}{n\|K_n\|_F},$$

where $Y = (y_1, \ldots, y_n)$, $\langle \cdot, \cdot\rangle_F$ is the Frobenius inner product and $\|\cdot\|_F$ is the Frobenius norm. In this section, we drop the subscript $F$. The following bound is suggested in (Jia & Liao, 2009),

$$P\{|A(K_n) - A(y)| > \epsilon\} \leq 2\exp\left(-\frac{2\epsilon^2(n-1)^2}{nC^2(\theta)}\right),$$

where $C(\theta) = |A(K_n)|\theta^{-1}(m - (m-1)\theta + \frac{2n-1}{\|K_n\|})$ and $\theta$ is defined by $\theta = 1 - \max_{s=1,\ldots,n} \min_{i=1,\ldots,n-1} \frac{\lambda_i(K_n^s)}{\lambda_i(K_n)}$, where $K_n^s$ is the submatrix of $K_n$, where the $s$-th row and column are replaced by $\mathbf{0}$ vectors. Here, we provide another concentration bound, which depends only on eigenvalues of the sample kernel matrix, and can be approximated by the ratio between the first and second largest eigenvalues of the kernel matrix.

**Theorem 4.1.** Let us take (Assumption 1), and $A(y)$ and $A(K)$ are defined as above. Then, we have

$$P\{|A(K_n) - A(y)| > \epsilon\}$$
$$\leq 2\exp\left(-\frac{2\epsilon^2}{A(K_n)\left|\frac{1}{n-1} - \frac{\|K_n\|}{L}\right| + \left(2 + \frac{1}{n-1}\right)\frac{1}{L}}\right).$$

**Proof.**(Sketch) The proof follows from the Cauchy interlacing lemma and the bounded difference inequality. Let us define $K_n^n$ as $K_n^s$ as above, where $s = n$. Then, we have

$$|A(K_n) - A(K_n^n)| \leq \left|\frac{\langle K_n, YY^\top\rangle}{n\|K_n\|} - \frac{\langle K_n^n, Y^n Y^{n,\top}\rangle}{(n-1)\|K_n^n\|}\right|$$
$$\leq \left|\frac{\langle K_n, YY^\top\rangle}{n\|K_n\|} - \frac{\langle K_n, YY^\top\rangle}{(n-1)\|K_n^n\|}\right|$$
$$+ \left|\frac{\langle K_n, YY^\top\rangle}{(n-1)\|K^n\|} - \frac{\langle K_n^n, Y^n Y^{n\top}\rangle}{(n-1)\|K_n^n\|}\right|$$
$$\leq |\langle K_n, YY^\top\rangle_F|\left|\frac{1}{(n-1)\|K_n^n\|} - \frac{1}{\|K_n\|}\right|$$
$$+ \frac{2n-1}{(n-1)\|K_n^n\|}.$$

The result follows by plugging in the following inequality into the above.

$$L := \sqrt{\sum_{i=2}^{n-1} \lambda_i^2(K_n)} \leq \|K_n^n\|_F \leq \sqrt{\sum_{i=1}^{n-1} \lambda_i^2(K_n)}.$$

Therefore, we have

$$|A(K_n) - A(K_n^n)|$$
$$= \frac{\langle K_n, YY^\top \rangle}{\|K_n\|} \left| \frac{(n-1)\|K_n\| - \|K_n^n\|}{(n-1)\|K_n^n\|} \right| + \frac{2n-1}{n-1} \frac{1}{\|K_n^n\|}$$
$$\leq A(K_n) \left| \frac{1}{n-1} - \frac{\|K_n\|}{L} \right| + \left(2 + \frac{1}{n-1}\right) \frac{1}{L}.$$
□

In the preceding results, the term $\frac{\|K_n\|}{L}$ can be approximated by $\frac{\lambda_1(K_n)}{\lambda_2(K_n)}$. Therefore, for sufficiently large samples size, the gap between first and second eigenvalues control the concentration of the sample kernel alignment.

## 5 CONCLUDING REMARKS

Kernel methods are very successful approaches in solving different machine learning problems. This success is mainly rooted in using feature maps and kernel functions, which transform the given samples into a possibly higher dimensional space.

Kernel matrices/operators can be characterized by the properties of their spectral decomposition. Also, in the computation of the excess risk, we eventually need the eigenvalue information. Concentration inequalities for the spectrum of kernel matrices measure how close the sample eigenvalues/eigenvectors are to the population values. The previous concentration results are either suboptimal, or computing the bound is impractical. This paper improves upon the optimality and computational feasibility of the previous results, by applying Cauchy's interlacing lemma. Moreover, for inner product and Euclidean distance kernels, e.g. radial basis function (Steinwart & Christmann, 2008), we derive new types of bounds, which connect the concentration of sample kernel eigenvalues and eigenvectors to the eigenvalues of the sample covariance matrix. This result may explain the poor performance of some nonlinear kernels in very high dimensions, as the largest eigenvalue of the sample covariance matrix become very large. As an interesting case study, we establish a computationally less demanding, simple and informative bound for the concentration of the sample kernel target-alignment.

The results can be improved, for example by using more careful calculations of the spectral norms in bounded difference computation, other operator perturbation results, or local concentration inequalities.

**Acknowledgements**

Authors acknowledge the reviewers' comments, which improved the paper considerably. NR and RV were funded by the Academy of Finland, through Centres of Excellence Program 2006-2011.